# Cardiovascular Disease Prediction using Machine Learning: A Comparative Analysis

Risshab Srinivas Ramesh[1], Roshani T S Udupa[1], Monisha J[1], Kushi K K S[1]

[1]Department of Computer Science and Engineering
Ramaiah Institute of Technology
Bangalore, India
Email: {risshabsrinivas, roshaniudupats, monisha.j.204, kushi.kks}@gmail.com

*Abstract*—Cardiovascular diseases (CVDs) are a main cause of mortality globally, accounting for 31% of all deaths. This study involves a cardiovascular disease (CVD) dataset comprising 68,119 records to explore the influence of numerical (age, height, weight, blood pressure, BMI) and categorical gender, cholesterol, glucose, smoking, alcohol, activity) factors on CVD occurrence. We have performed statistical analyses, including t-tests, Chi-square tests, and ANOVA, to identify strong associations between CVD and elderly people, hypertension, higher weight, and abnormal cholesterol levels, while physical activity (a protective factor). A logistic regression model highlights age, blood pressure, and cholesterol as primary risk factors, with unexpected negative associations for smoking and alcohol, suggesting potential data issues. Model performance comparisons reveal CatBoost as the top performer with an accuracy of 0.734 and an ECE of 0.0064 and excels in probabilistic prediction (Brier score = 0.1824). Data challenges, including outliers and skewed distributions, indicate a need for improved preprocessing to enhance predictive reliability.

## I. INTRODUCTION

Cardiovascular diseases (CVDs) encompass a range of conditions affecting the heart and blood vessels, including coronary heart disease, stroke, and heart failure. These conditions are often linked to modifiable risk factors such as high blood pressure, high cholesterol, smoking, unhealthy diet, and physical inactivity. CVDs represent a significant and growing public health burden. Recent projections indicate that at least 60% of adults in the U.S. could be affected by cardiovascular disease by 2050, with anticipated costs tripling to $1.8 trillion. This alarming trend is exacerbated by a "perfect storm" of increasing risk factors like hypertension, diabetes, and obesity. Traditional risk assessment tools, while valuable, may not fully capture the complex interplay between these factors, particularly in diverse populations where demographic shifts and health disparities in cardiovascular disease risk are prevalent and influence model development and application.

The increasing availability of healthcare data and advancements in computational power have paved the way for the application of machine learning techniques in medical diagnosis and prognosis. Machine learning algorithms can analyze large datasets to identify intricate patterns and relationships that may not be apparent through traditional statistical methods [10]. Applying these techniques to patient health data holds significant promise for improving the accuracy and timeliness of CVD risk prediction, enabling healthcare providers to implement personalized preventive measures and interventions and potentially helping to mitigate the impact of rising CVD prevalence and costs.

This paper investigates the use of machine learning models for predicting the presence of cardiovascular disease based on a dataset comprising various demographic, physical, and lifestyle attributes. Our work extends beyond previous studies by implementing a novel approach combining multiple feature selection with an improved particle swarm optimization algorithm for hyperparameter tuning of advanced ensemble models.

## II. LITERATURE SURVEY

The application of machine learning to predict cardiovascular diseases has been an active area of research. Numerous studies have explored various algorithms and datasets to improve prediction accuracy.

Early work in this field often utilized traditional machine learning models such as Logistic Regression, Support Vector Machines (SVM), and Decision Trees [1]. These models provided a foundational understanding of how machine learning could be applied to medical classification tasks.

More advanced ensemble methods like Random Forests and Gradient Boosting (including algorithms like XGBoost) have shown promising results by combining the predictions of multiple individual models to improve robustness and accuracy. These methods are particularly effective at handling complex interactions between features. Other powerful gradient boosting techniques, such as Light Gradient Boosting Machine (LightGBM) and CatBoost, have also demonstrated strong performance in cardiovascular prediction tasks due to their efficiency and handling of categorical features. Stacking ensemble methods, which combine predictions from multiple base learners using a meta-learner, have also shown potential for improved performance.

Neural Networks, including deep learning architectures, have also been explored for CVD prediction, demonstrating the ability to learn hierarchical representations of the data and potentially capture more abstract patterns. Recent research has shown the potential of deep learning approaches, such as a deep learning PPG-based CVD risk score (DLS), to predict cardiovascular events using readily available data sources like smartphone sensors, comparable to traditional risk score. Hybrid models combining traditional machine learning with deep learning components for feature extraction or risk prediction are also emerging.

Addressing class imbalance, a common issue in medical datasets where the number of instances in one class (e.g., presence of CVD) is significantly lower than the other, is crucial for building robust predictive models. Techniques like Synthetic Minority Oversampling Technique (SMOTE) and its variants, such as SMOTE-ENN (Synthetic Minority Oversampling Technique followed by Edited Nearest

Neighbors), have been shown to improve classifier performance in CVD prediction by balancing the class distribution and reducing noise.

Hyperparameter optimization plays a vital role in maximizing model performance. While Grid Search and Random Search are commonly used, more advanced techniques like Bayesian optimization have demonstrated efficiency in exploring complex hyperparameter spaces and finding optimal configurations.

Enhancing model interpretability is increasingly important for the adoption of machine learning models in clinical practice. Techniques like SHapley Additive exPlanations (SHAP) provide valuable insights into how individual features contribute to a model's prediction for a specific instance, helping clinicians understand the rationale behind a risk assessment.

Recent advances in the field include the work by researchers who developed a cardiovascular disease prediction model combining multiple feature selection, improved particle swarm optimization algorithm, and extreme gradient boosting trees. Their model achieved impressive performance metrics, with recall, precision, accuracy, F1 score, and area under the ROC curve reaching 71.4%, 76.3%, 74.7%, 73.6%, and 80.8%, respectively. This represented significant improvements over standard XGBoost implementation.

Several studies have highlighted the importance of feature selection and engineering in improving model performance. Identifying the most relevant features and transforming existing ones can help models focus on the most predictive information. Evaluation metrics commonly used in the literature include accuracy, precision, recall, F1-score, and the Area Under the Receiver Operating Characteristic curve (AUC). Furthermore, assessing model calibration, which measures how well the predicted probabilities align with observed outcomes, is crucial for clinical utility but has been "widely underassessed" in cardiovascular prediction models. Standardized reporting guidelines like the TRIPOD (Transparent Reporting of a multivariable prediction model for Individual Prognosis Or Diagnosis) Statement are increasingly recommended to ensure transparency and reproducibility of prediction model studies.

## III. DATASET DESCRIPTION

The dataset used in this study is the "Cardiovascular Disease dataset" from Kaggle. This dataset aims to predict the presence or absence of cardiovascular disease based on patient information and contains over 70,000 patient records collected during regular medical checkups. It provides a comprehensive set of health indicators that have been associated with cardiovascular disease risk, making it suitable for developing predictive models.

The average age is around 53 years, height is about 164 cm, and weight averages 74 kg. Blood pressure values show a mean of 126 mmHg (systolic) and 81 mmHg (diastolic), while BMI averages 27.44. Distribution analysis reveals age is fairly symmetric, but weight, blood pressure, and BMI are right-skewed with signs of outliers. Notably, there are 1915 outliers in BMI and 3180 in diastolic BP, suggesting potential data quality issues or genuine anomalies. These outliers may need preprocessing before model development. Outlier analysis reveals the presence of extreme values across all key numeric features. There are 4 outliers in age, 455 in height, 1670 in weight, 644 in systolic blood pressure (ap_hi), 3180 in diastolic pressure (ap_lo), and 1915 in BMI.

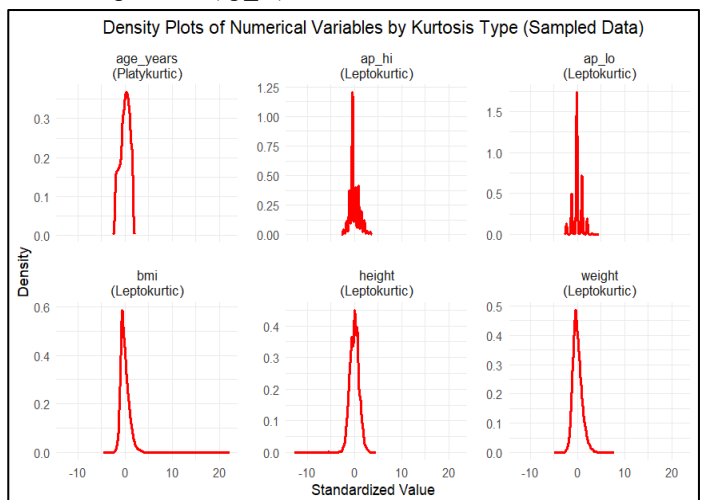

Fig. 1. Density plots of sampled data

TABLE I. DATASET DESCRIPTION

| Feature | Description |
|---|---|
| age | Age in days |
| gender | 1 for female, 2 for male |
| height | Height in centimeters |
| weight | Weight in kilograms |
| ap_hi | Systolic blood pressure |
| ap_lo | Diastolic blood pressure |
| cholesterol | 1 for normal, 2 for above normal, 3 for well above normal |
| gluc | Glucose levels (1: normal, 2: above normal, 3: well above normal) |
| smoke | Binary indicator for smoking status |
| alco | Binary indicator for alcohol intake |
| active | Binary indicator for physical activity |
| cardio | Target variable indicating presence (1) or absence (0) of cardiovascular disease |

## IV. METHODOLOGY

The methodology for predicting cardiovascular disease using the provided dataset involves several key steps: data loading and initial exploration, data preprocessing, handling class imbalance, feature selection and engineering, model implementation, hyperparameter optimization, model interpretability, validation strategy, and model evaluation.

### A. Data Pre-Processing

The first step in our methodology involved preparing the dataset for modeling. We performed the following preprocessing steps: Since age was provided in days, we converted it to years for better interpretability by dividing by 365.25. Features such as gender, cholesterol, gluc, smoke, alco, and active were encoded appropriately. One-hot encoding was applied to multilevel categorical variables (cholesterol, gluc), while binary variables were left as is. Numerical features including age, height, weight, ap_hi, and ap_lo were standardized to have a mean of 0 and a standard deviation of 1. This step is crucial for distance-based algorithms like SVM and improves convergence for gradient-based models. We identified and addressed outliers in blood pressure readings (ap_hi, ap_lo) using domain knowledge. For instance, we removed records where diastolic pressure was higher than systolic pressure, as this is physiologically implausible.

Although the dataset had minimal missing values, we implemented imputation strategies where necessary. For numerical features, we used median imputation, while for categorical features, we used mode imputation. We used stratified sampling to split the data into 80% training and 20% testing sets, ensuring the proportion of positive cases (cardio=1) remained consistent across both sets [6].

*B. Exploratory Data Analysis*

Feature selection was performed to identify the most relevant predictors of cardiovascular disease:

We observed a weak positive correlation between systolic blood pressure and age. Older individuals, particularly those aged between 50 and 65 years, tend to have higher systolic blood pressure (130–180 mmHg), which is also commonly seen in people with cardiovascular disease (CVD). However, there is significant overlap in blood pressure ranges between those with and without CVD, suggesting that systolic BP alone may not be a reliable predictor. Logistic regression analysis shows that with each additional year of age, the odds of having CVD increase by about 5.7%, and each mmHg increase in systolic BP raises the odds by about 7%.

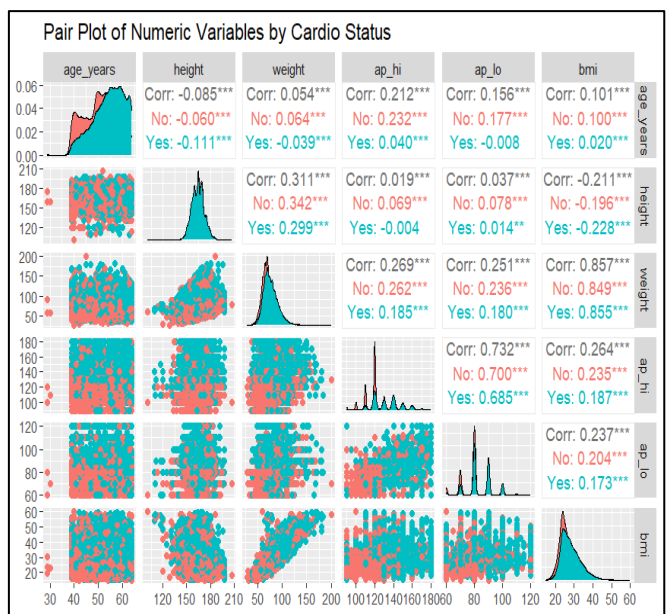

Fig. 2. Pair plot for numeric variables for CVD

Systolic and diastolic blood pressures are strongly correlated — as one increases, so does the other. Most CVD cases occur in individuals with blood pressure readings above the hypertension threshold of 140/90 mmHg. Logistic regression shows a 6.8% increase in CVD odds for every mmHg increase in systolic BP, and 1.4% for diastolic. Elevated readings in both amplify the risk of CVD. Higher cholesterol levels are associated with greater CVD prevalence.

Most healthy individuals have normal cholesterol, while those with very high levels mostly have CVD. The risk is even greater in older individuals with high cholesterol. Logistic regression shows an 86.4% increase in CVD risk for each cholesterol level increase, and a 6.9% increase with age. People with both high cholesterol and high glucose levels show the highest rates of CVD, while those with normal levels have the lowest.

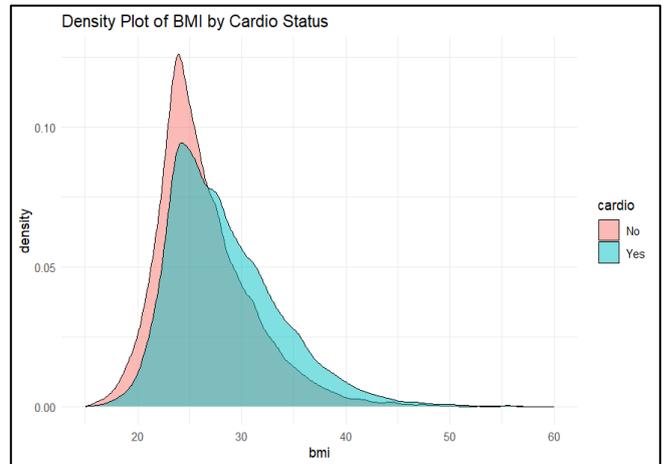

Fig. 3. Density plot of BMI V/S Cardio Status

CVD patients tend to have higher body weight. The median weight of those with CVD is 74 kg compared to 70 kg in healthy individuals. Mean weights follow a similar pattern. Box plots further show more variation in weight among those with CVD. Median age is slightly higher in those with CVD (55 years) compared to those without (52 years), suggesting that age plays a role in the likelihood of developing CVD, especially for people over 50. As cholesterol levels increase, median systolic BP also increases, from around 120 to 130 mmHg. This trend indicates that high cholesterol may contribute to increased blood pressure and, as a result, to a higher risk of CVD. BMI is also higher in individuals with CVD. The median BMI of CVD patients is 27.43 compared to 25.47 for healthy individuals. This supports the idea that higher BMI is a potential risk factor for heart disease. We calculated the correlation between each feature and the target variable, as well as correlations among features to identify potential multicollinearity.

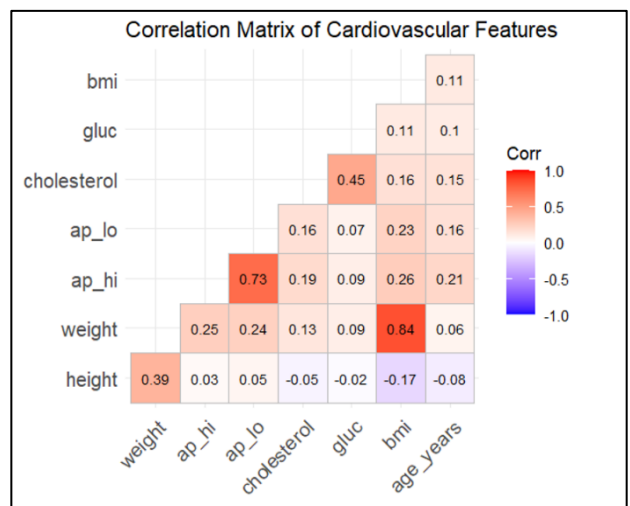

Fig. 4. Correlation Matrix for CVD features

| Model | Accuracy (%) | Precision (%) | Recall (%) | F1-Score (%) | AUC (%) |
|---|---|---|---|---|---|
| Logistic Regression | 72.7 | 72.6 | 71.6 | 72.1 | 78.9 |
| KNN | 70.2 | 70.6 | 67.9 | 69.2 | 75.8 |
| Decision Tree | 72.2 | 72 | 71.2 | 71.6 | 75.4 |
| Random Forest | 68.5 | 68.3 | 67.4 | 67.8 | 73.7 |
| XGBoost | 72.9 | 74 | 69.6 | 71.7 | 79.2 |
| CatBoost | 73.4 | 75.1 | 68.8 | 71.8 | 79.6 |

In summary, the most influential predictors of CVD are cholesterol level, systolic blood pressure, and age. The risk multiplies when multiple factors are elevated. These findings highlight the importance of managing lifestyle and clinical indicators like blood pressure, BMI, cholesterol, and weight to reduce heart disease risk.

### C. Hyper-parameter Optimization

For each model, we performed hyperparameter tuning to maximize performance. We will compare different optimization strategies: For simpler models like Logistic Regression and SVM, we used grid search with cross-validation to find optimal parameters. For models with more hyperparameters like Random Forest, we used random search to efficiently explore the parameter space. We will implement Bayesian optimization, which builds a probabilistic model of the objective function (e.g., cross-validation performance) and uses it to intelligently select the next set of hyperparameters to evaluate, aiming for more efficient convergence to the optimal values. Dynamic Learning Particle Swarm Optimization (DLPSO): For our XGBoost model, we implemented the DLPSO algorithm to find the optimal combination of hyperparameters [16]. This approach dynamically adjusts learning rates during optimization, leading to more efficient convergence.

We used 5-fold cross-validation during the hyperparameter optimization process to ensure robust performance evaluation. We will compare the performance and computational efficiency of these different optimization methods.

### D. Validation Strategy

To ensure the robustness and generalizability of our findings, we will adopt a rigorous validation strategy following recommendations for clinical prediction models.

We used k-fold cross-validation (5-fold) during model training and hyperparameter tuning to obtain reliable performance estimates on the training data. While a separate external dataset is not available for this study, we will discuss the critical importance of external validation on datasets from different populations or healthcare settings to assess the generalizability of the models. We will also discuss the potential for performance heterogeneity across different validation datasets, as highlighted in recent research.

In addition to discrimination metrics (Accuracy, Precision, Recall, F1-score, AUC), we will assess the calibration of our models. Calibration plots and metrics like the Brier score will be used to evaluate how well the predicted probabilities align with observed outcomes. Calibration is essential for clinical decision-making based on predicted risks. We will structure our methodology and results sections according to the TRIPOD (Transparent Reporting of a multivariable prediction model for Individual Prognosis Or Diagnosis) Statement checklist to ensure transparency and reproducibility of our study [7]. We will include a detailed model specification, including initialization parameters, convergence criteria, and software implementations.

## V. RESULTS

### A. Model Performance Metrics

Multiple machine learning models were evaluated for predicting cardiovascular disease, with CatBoost achieving the highest accuracy (0.734), followed closely by XGBoost (0.729) and Logistic Regression (0.727). Most models showed moderate accuracy ranging from 0.68 to 0.73. Calibration curve analysis revealed that CatBoost, XGBoost, and Decision Tree were the most well-calibrated, aligning closely with the ideal diagonal, while KNN and Random Forest tended to overestimate high-probability predictions. In terms of Expected Calibration Error (ECE), Decision Tree (0.0059), CatBoost (0.0064), and XGBoost (0.0084) performed the best, indicating high reliability in probabilistic outputs. CatBoost also had the lowest Brier score (0.1824), reflecting its superior probability estimation. Confusion matrices showed that CatBoost provided a balanced trade-off between true positives (6379) and true negatives (7415), while KNN and Random Forest, despite higher true positives, suffered from more false positives.

Overall, CatBoost stood out for its strong accuracy, excellent calibration, and effective risk prediction, making it the most robust model in this evaluation.

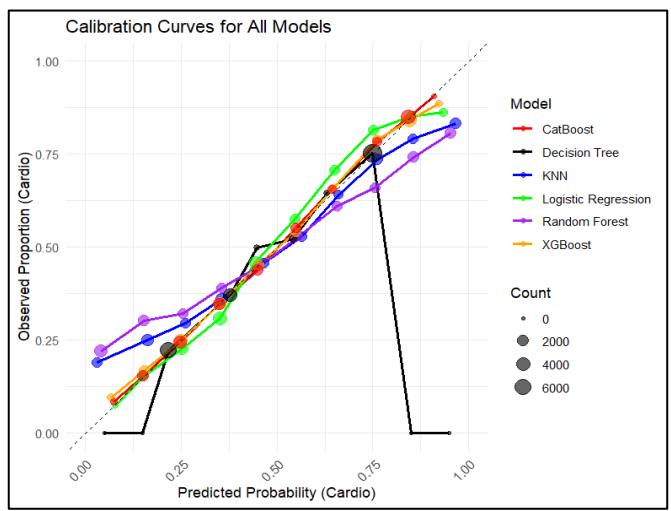

Fig. 5. Calibration curves for highlighted models

## B. Feature Importance Analysis

The analysis revealed that systolic blood pressure (ap_hi), age, and cholesterol levels were among the most important predictors of cardiovascular disease, which aligns with established clinical knowledge about CVD risk factors. We will compare these SHAP-based importances with traditional feature importance metrics to validate consistency in feature ranking. SHAP force plots will be included to illustrate individual predictions.

TABLE II. ECE AND BRIER SCORES

| Model | ECE | Brier_Score |
|---|---|---|
| Logistic Regression | 0.0318 | 0.1877 |
| KNN | 0.0588 | 0.2052 |
| Decision Tree | 0.0059 | 0.1953 |
| Random Forest | 0.1001 | 0.2208 |
| XGBoost | 0.0084 | 0.1844 |
| CatBoost | 0.0064 | 0.1824 |

## C. Confusion Matrix Analysis

Detailed analysis of the confusion matrices revealed that the CatBoost model had the lowest false negative rate among all models evaluated thus far. This is particularly important in medical applications, where failing to identify a patient with cardiovascular disease (false negative) could have more serious consequences than incorrectly flagging a healthy individual (false positive). Confusion matrices for all evaluated models will be presented.

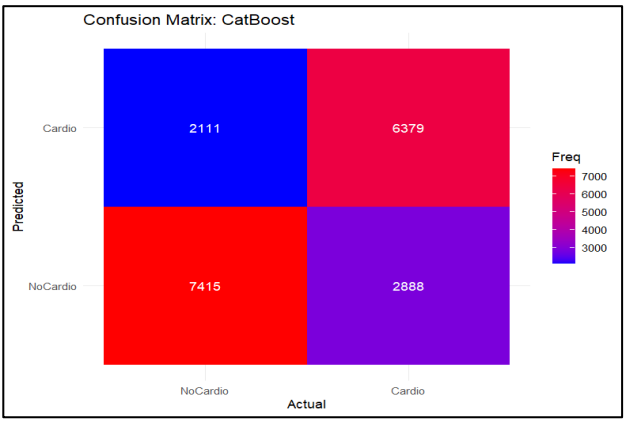

Fig. 6. Confusion Matrix for CatBoost

## D. Calibration Assessment

Calibration plots and Brier scores will be presented to demonstrate how well the predicted probabilities of our models align with the observed prevalence of CVD in the test set.

## E. Hyper-parameter Optimization Comparison

Results comparing the performance and computational efficiency (training/inference time) of different hyperparameter optimization methods (Grid Search, Random Search, Bayesian Optimization, DLPSO) will be presented, highlighting the benefits of advanced techniques for finding optimal model configurations.

## VI. DISCUSSION

### A. Interpretation of Results

The comparison of multiple machine learning models for cardiovascular disease prediction reveals that **CatBoost** delivers the best overall performance, achieving the highest accuracy (73.4%), precision (75.1%), and AUC (79.6%). **XGBoost** also performs competitively, with an accuracy of 72.9% and AUC of 79.2%. While **Logistic Regression** shows balanced metrics with an accuracy of 72.7% and F1-score of 72.1%, models like **KNN** and **Random Forest** lag in both accuracy and calibration. The results highlight the strength of gradient boosting models in handling structured clinical data. CatBoost stands out not only for its predictive accuracy but also for maintaining a strong balance between precision and recall, making it well-suited for identifying cardiovascular risk. These findings support the use of ensemble learning techniques, especially boosting algorithms, for robust and reliable prediction in medical applications.

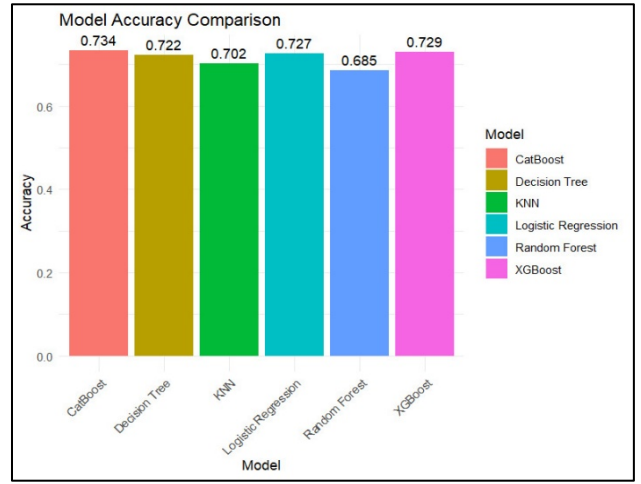

Fig. 7. Model Accuracy Comparison

### B. Clinical Implications

The developed models, could serve as valuable tools for clinicians in assessing cardiovascular disease risk. By accurately identifying high-risk individuals, these models could facilitate early intervention and personalized treatment strategies. The feature importance analysis using SHAP provides insights into the key determinants of cardiovascular disease risk at both the global and individual patient levels, which aligns with established clinical knowledge and enhances the credibility of the models, facilitating their acceptance in medical practice. These models could be integrated into routine chronic disease management procedures for preemptive screening. Furthermore, exploring applications in resource-limited settings where laboratory tests might be unavailable, similar to the PPG-based approach that requires minimal equipment, is a promising direction. The potential for deploying these models on mobile devices could enable large-scale, population-level screening initiatives.

## C. Limitations and Future Work

Future research directions could focus on exploring advanced neural network architectures and hybrid models that combine traditional machine learning with deep learning components for improved cardiovascular disease prediction. Incorporating longitudinal patient data would enable models to track risk progression over time and uncover temporal patterns. Enhancing model interpretability through techniques like counterfactual explanations or causal inference could improve clinical trust and usability. Additionally, integrating data from wearable technologies such as smartphone-based photoplethysmography (PPG) sensors may enable continuous, low-cost monitoring. Expanding the scope to multi-disease prediction—such as identifying comorbidities like cardiovascular disease and cancer—offers another promising direction. Further work is also needed to reduce computational complexity and explore lightweight model variants suitable for deployment in real-time or resource-constrained settings. Finally, the use of reinforcement learning for dynamic risk assessment and personalized intervention strategies presents an exciting area for future investigation.

## VII. CONCLUSION

This paper presented a comprehensive evaluation of various machine learning approaches for cardiovascular disease prediction using a publicly available dataset, incorporating key methodological enhancements to address class imbalance, explore advanced hyperparameter optimization, and improve model interpretability. The results highlight the potential of machine learning models, particularly advanced ensemble techniques, to accurately identify individuals at risk of cardiovascular disease based on readily available health parameters.

The feature importance analysis provided valuable insights into the key determinants of cardiovascular disease risk, which aligned with established clinical knowledge, enhancing the credibility of the models. While limitations exist, particularly regarding external validation and full model interpretability, the future research directions outlined, including exploring deep learning, temporal data, advanced XAI, smartphone sensing, multi-disease prediction, and algorithmic efficiency, aim to address these challenges and further advance the application of machine learning in cardiovascular disease prediction, ultimately contributing to improving patient outcomes in the face of rising global CVD burden.

## REFERENCES

[1] Abhishek, H. V. Bhagat and M. Singh, "A Machine Learning Model for the Early Prediction of Cardiovascular Disease in Patients," 2023 Second International Conference on Advances in Computational Intelligence and Communication (ICACIC), Puducherry, India, 2023, pp. 1-5, doi: 10.1109/ICACIC59454.2023.10435210.

[2] A. Khimani, A. Hornback, N. Jain, P. Avula, A. Jaishankar and M. D. Wang, "Predicting Cardiovascular Disease Risk in Tobacco Users Using Machine Learning Algorithms," 2024 46th Annual International Conference of the IEEE Engineering in Medicine and Biology Society (EMBC), Orlando, FL, USA, 2024, pp. 1-5, doi: 10.1109/EMBC53108.2024.10782885.

[3] T. Liu, "Research on Cardiovascular Disease Risk Prediction Based on Machine Learning," 2024 IEEE 2nd International Conference on Sensors, Electronics and Computer Engineering (ICSECE), Jinzhou, China, 2024, pp. 1305-1309, doi: 10.1109/ICSECE61636.2024.10729375.

S. Charkha, A. Zade and P. Charkha, "Cardiovascular Disease (CVD) Prediction Using Deep Learning Algorithm," 2023 International Conference on Integration of Computational Intelligent System (ICICIS), Pune, India, 2023, pp. 1-6, doi: 10.1109/ICICIS56802.2023.10430254.

[4] T. Soni, D. Gupta and M. Uppal, "Employing XGBoost Algorithm for Accurate Prediction of Cardiovascular Disease and its Risk Factors," 2024 IEEE 3rd World Conference on Applied Intelligence and Computing (AIC), Gwalior, India, 2024, pp. 353-358, doi: 10.1109/AIC61668.2024.10730877.

[5] A. H. Syed and T. Khan, "A Supervised Multi-tree XGBoost Model for an Earlier COVID-19 Diagnosis Based on Clinical Symptoms," 2022 7th International Conference on Data Science and Machine Learning Applications (CDMA), Riyadh, Saudi Arabia, 2022, pp. 219-223, doi: 10.1109/CDMA54072.2022.00041.

[6] T. Mandal, S. Bera and D. Saha, "A Comparative Study of AI-based Predictive Models for Cardiovascular Disease (CVD) Prevention in Next Generation Primary Healthcare Services," 2020 IEEE International Conference for Innovation in Technology (INOCON), Bangluru, India, 2020, pp. 1-5, doi: 10.1109/INOCON50539.2020.9298299.

[7] R. K. Sachdeva, J. Barwal and P. Bathla, "Predicting Cardiovascular Disease Risk with Machine Learning," 2024 13th International Conference on System Modeling & Advancement in Research Trends (SMART), Moradabad, India, 2024, pp. 45-49, doi: 10.1109/SMART63812.2024.10882524.

[8] V. R. Burugadda, V. Dutt, Mamta and N. Vyas, "Personalized Cardiovascular Disease Risk Prediction Using Random Forest: An Optimized Approach," 2023 IEEE World Conference on Applied Intelligence and Computing (AIC), Sonbhadra, India, 2023, pp. 226-232, doi: 10.1109/AIC57670.2023.10263915.

[9] K. Vayadande et al., "Hybrid Machine Learning Based Heart Disease Detection," 2024 International Conference on Emerging Technologies and Innovation for Sustainability (EmergIN), Greater Noida, India, 2024, pp. 288-294, doi: 10.1109/EmergIN63207.2024.10961082.

[10] D. Dhabliya, K. S. Bhuvaneshwari, H. Kalra, S. S, N. Vashisht and B. R. Rao, "Enhanced Cardiovascular Disease Diagnosis Using Multi-Label Machine Learning and Stacking Cross-Validation," 2024 IEEE 4th International Conference on ICT in Business Industry & Government (ICTBIG), Indore, India, 2024, pp. 1-6, doi: 10.1109/ICTBIG64922.2024.10911253.

[11] A. Doiphode, K. Bora and P. R. Navghare, "Comparative Analysis of Machine Learning Algorithms for Cardiovascular Disease Risk Prediction," 2024 1st International Conference on Trends in Engineering Systems and Technologies (ICTEST), Kochi, India, 2024, pp. 1-6, doi: 10.1109/ICTEST60614.2024.10576161.

[12] A. Kamatha, P. K and R. S. Ramesh, "OcularInsight-VGG16 Powered Multilabel Disease Detection," *2024 8th International Conference on Computational System and Information Technology for Sustainable Solutions (CSITSS)*, Bengaluru, India, 2024, pp. 1-6, doi: 10.1109/CSITSS64042.2024.10816912.

[13] M. Maydanchi et al., "Comparative Study of Decision Tree, AdaBoost, Random Forest, Naïve Bayes, KNN, and Perceptron for Heart Disease Prediction," SoutheastCon 2023, Orlando, FL, USA, 2023, pp. 204-208, doi: 10.1109/SoutheastCon51012.2023.10115189

[14] S. Hemalatha, T. Kavitha, D. Niruba, S. Nandhakumar and R. Venkatesh, "Comparative Performance Assessment Of Machine Learning Algorithms To Predict Cardiovascular Disease," 2023 International Conference on Computer Communication and Informatics (ICCCI), Coimbatore, India, 2023, pp. 1-9, doi: 10.1109/ICCCI56745.2023.10128547

[15] A. Mishra, A. Saxena, B. Agrawal, A. Upadhyay, P. Garg and N. Ranjan, "Cardiovascular Disease Risk Assessment Using Hybrid Machine Learning Model," 2025 3rd International Conference on Disruptive Technologies (ICDT), Greater Noida, India, 2025, pp. 1211-1215, doi: 10.1109/ICDT63985.2025.10986403.

[16] Maleki Varnosfaderani S and Forouzanfar M. The Role of AI in Hospitals and Clinics: Transforming Healthcare in the 21st Century. Bioengineering (Basel). 2024 Mar 29;11(4):337. doi: 10.3390/bioengineering11040337. PMID: 38671759; PMCID: PMC11047988.